\documentclass{article} 

\usepackage{eso-pic, rotating, graphicx}
\AddToShipoutPicture{\put(550,50){\rotatebox{90}{
\resizebox{22cm}{!}{%
  \begin{tabular}{@{}l@{}}
        This is the author's version.
        This paper was published in Frontiers in Neurorobotics:\\
        \url{www.frontiersin.org/articles/10.3389/fnbot.2019.00081}
  \end{tabular}%
}}}}

\usepackage{amsbsy}
\usepackage{amsmath}
\usepackage[margin=1.0in]{geometry}

\usepackage{url,hyperref,lineno,microtype,subcaption}
\usepackage[onehalfspacing]{setspace}

\usepackage[utf8]{inputenc}

\usepackage{todonotes}
\usepackage{csquotes}
\usepackage{comment}
\usepackage[noabbrev,capitalize]{cleveref}
\usepackage{gensymb}
\usepackage{acronym}
\usepackage{siunitx}
\usepackage{booktabs}

\acrodef{SPORE}[SPORE]{Synaptic Plasticity with Online REinforcement learning}
\acrodef{PPO}[PPO]{Proximal Policy Optimization}
\acrodef{RBM}[RBM]{Restricted Boltzmann Machine}

\newenvironment{params}[1]{%
\noindent
\begin{table}[h]
  \caption{#1}
  \vspace{0.2cm}
\begin{tabular}{@{\hspace{2em}}ll}
  \toprule
}{%
\bottomrule
\end{tabular}
\end{table}%
}

\newcommand{\normal}[1]{\mathcal{N}(#1)}

\newcommand{\ve}[1]{\boldsymbol{#1}}

\newcommand{\bth}{\ve \theta}

\newcommand{\ddthetai}{\frac{\partial}{\partial \theta_i}}

\newcommand{\ps}[1]{p\left({#1}\right)}

\newcommand{\wiener}{\mathcal{W}}
\newcommand{\expect}[2][]{\left\langle\, #2\, \right\rangle_{#1} }

\newcommand{\rtau}{r(\tau)}
\newcommand{\rseq}{\ve r}

\title{Embodied Synaptic Plasticity with Online Reinforcement learning}

\author{
  Jacques Kaiser$^{1}$,
  Michael Hoff$^{1,2}$,
  Andreas Konle$^1$,
  J. Camilo Vasquez Tieck$^1$,\\
  David Kappel$^{2,3,4}$,
  Daniel Reichard$^1$,
  Anand Subramoney$^2$,
  Robert Legenstein$^2$,\\
  Arne Roennau$^1$,
  Wolfgang Maass$^2$,
  R\"udiger Dillmann$^1$\\
  {\small $^{1}$FZI Research Center for Information Technology,}\\
  {\small $^{2}$Graz University of Technology,}\\
  {\small $^3$ Georg-August Universit\"at, G\"ottingen,}\\
  {\small $^4$ Technische Universit\"at Dresden}
}

\begin{document}

\maketitle

\begin{abstract}
  The endeavor to understand the brain involves multiple collaborating research fields.
  Classically, synaptic plasticity rules derived by theoretical neuroscientists are evaluated in isolation on pattern classification tasks.
  This contrasts with the biological brain which purpose is to control a body in closed-loop.
  This paper contributes to bringing the fields of computational neuroscience and robotics closer together by integrating open-source software components from these two fields.
  The resulting framework allows to evaluate the validity of biologically-plausibe plasticity models in closed-loop robotics environments.
  We demonstrate this framework to evaluate \ac{SPORE}, a reward-learning rule based on synaptic sampling, on two visuomotor tasks: reaching and lane following.
  We show that \ac{SPORE} is capable of learning to perform policies within the course of simulated hours for both tasks.
  Provisional parameter explorations indicate that the learning rate and the temperature driving the stochastic processes that govern synaptic learning dynamics need to be regulated for performance improvements to be retained.
  We conclude by discussing the recent deep reinforcement learning techniques which would be beneficial to increase the functionality of \ac{SPORE} on visuomotor tasks.
\end{abstract}

\acresetall 

\section{Introduction}



The brain evolved over millions of years for the sole purpose of controlling the body in a goal-directed fashion.
Computations are performed relying on neural dynamics and asynchronous communication.
Spiking neural network models base their computations on these computational principles.
Biologically plausible synaptic plasticity rules for functional learning in spiking neural networks are regularly proposed (\cite{zenke2018superspike,kaiser2018synaptic,Neftci17_stocsyna,Pfister_etal06_optispik,Urbanczik_Senn14_learby}).
In general, these rules are derived to minimize a distance (referred to as error) between the output of the network and a target.
Therefore, the evaluation of these rules is usually carried out on open-loop pattern classification tasks.
By neglecting the embodiment, this type of evaluation disregards the closed-loop dynamics the brain has to handle with the environment.
Indeed, the decisions taken by the brain have an impact on the environment, and this change is sensed back by the brain.
To get a deeper understanding of the plausibility of these rules, an embodied evaluation is necessary.
This evaluation is technically complicated since spiking neurons are dynamical systems that must be synchronized with the environment.
Additionally, as in biological bodies, sensory information and motor commands need to be encoded and decoded respectively.

In this paper, we bring the fields of computational neuroscience and robotics closer together by integrating open-source software components from these two fields.
The resulting framework is capable of learning online the control of simulated and real robots with a spiking network in a modular fashion.
This framework is demonstrated in the evaluation of the promising neural reward-learning rule \ac{SPORE}~(\cite{Kappel2018,Kappel2015,Kappel2014,Yu2016a}) on two closed-loop robotic tasks.
\ac{SPORE} is an instantiation of the synaptic sampling scheme introduced in \cite{Kappel2018,Kappel2015}.
It incorporates a policy sampling method which models the growth of dendritic spines with respect to dopamine influx.
Unlike current state-of-the-art reinforcement learning methods implemented with conventional neural networks (\cite{mnih2015human,mnih2016asynchronous,lillicrap2015continuous}),
\ac{SPORE} learns online from precise spike-time and is entirely implemented with spiking neurons.
We evaluate this learning rule in a closed-loop reaching and a lane following~(\cite{bing2018end,kaiser2016towards}) setup.
In both tasks, an end-to-end visuomotor policy is learned, mapping visual input to motor commands.
In the last years, important progress have been made on learning control from visual input with deep learning.
However, deep learning approaches are computationally expensive and rely on biologically implausible mechanisms such as dense synchronous communication and batch learning.
For networks of spiking neurons learning visuomotor tasks online with synaptic plasticity rules remains challenging.
In this paper, visual input is encoded in Address Event Representation with a Dynamic Vision Sensor (DVS) simulation~(\cite{Lichtsteiner2008,kaiser2016towards}).
This representation drastically reduces the redundancy of the visual input as only motion is sensed, allowing more efficient learning.
It agrees with the two pathways hypothesis which states that motion is processed separately than color and shape in the visual cortex (\cite{Kruger2013}).


The main contribution of this paper is the embodiment of \ac{SPORE} and its evaluation on two neurorobotic tasks using a combination of open-source software components.
This embodiment allowed us to identify crucial techniques to regulate \ac{SPORE} learning dynamics,
not discussed in previous works where this learning rule was only evaluated on simple proof-of-concept learning problems~(\cite{Kappel2018,Kappel2015,Kappel2014,Yu2016a}).
Our results suggest that an external mechanism such as learning rate annealing is beneficial to retain a performing policy on advanced lane following task.

This paper is structured as follows.
We provide a review of the related work in \cref{sec:related}.
In \cref{sec:method}, we give a brief overview of \ac{SPORE} and discuss the contributed techniques required for its embodiment.
The implementation and evaluation on the two chosen neurorobotic tasks is carried out in \cref{sec:eval}.
Finally, we discuss in \cref{sec:conclusion} how the method could be improved.

\section{Related Work}
\label{sec:related}

The year 2015 marked a significant breakthrough in deep reinforcement learning.
Artificial neural networks of analog neurons are now capable of solving a variety of tasks ranging from  playing video games (\cite{mnih2015human}), to controlling multi-joints robots (\cite{Schulman2017,lillicrap2015continuous}) and lane following (\cite{Wolf2017}).
Most recent methods (\cite{Schulman2017,schulman2015trust,lillicrap2015continuous,mnih2016asynchronous}) are based on policy-gradients.
Specifically, policy parameters are updated by performing ascending gradient steps with backpropagation to maximize the probability of taking rewarding actions.
While functional, these methods are not based on biologically plausible processes.
First, a large part of neural dynamics are ignored.
Importantly, unlike \ac{SPORE}, these methods do not learn online -- weight updates are performed with respect to entire trajectories stored in rollout memory.
Second, learning is based on backpropagation which is not biologically plausible learning mechanism, as stated in \cite{bengio2015towards}.

Spiking network models inspired by deep reinforcement learning techniques were introduced in \cite{tieck2018learning} and \cite{Bellec2018}.
In both papers, the spiking networks are implemented with deep learning frameworks (PyTorch and TensorFlow, respectively) and rely on automatic differentiation.
Their policy-gradient approach is based on \ac{PPO} (\cite{Schulman2017}).
As the learning mechanism consists of backpropagating the \ac{PPO} loss (through-time in the case of \cite{Bellec2018}), most biological constraints stated in \cite{bengio2015towards} are still violated.
Indeed, the computations are based on spikes (4), but the backpropagation is purely linear (1), the feedback paths require precise knowledge of the derivatives (2) and weights (3) of the corresponding feedforward paths, and the feedforward and feedback phases alternate synchronously (5) (the enumeration refers to \cite{bengio2015towards}).

Only a small body of work focused on reinforcement learning with spiking neural networks, while addressing the previous points.
Groundwork of reinforcement learning with spiking networks was presented in \cite{izhikevich2007solving,florian2007reinforcement,legenstein2008learning}.
In these works, a mathematical formalization is introduced characterizing how dopamine modulated spike-timing-dependent plasticity (DA-STDP) solves the distal reward problem with eligibility traces.
Specifically, since the reward is received only after a rewarding action is performed, the brain needs a form of memory to reinforce previously chosen actions.
This problem is solved with the introduction eligibility traces, which assign credit to recently active synapses.
This concept has been observed in the brain (\cite{frey1997synaptic,pan2005dopamine}), and \ac{SPORE} also relies on eligibility traces.
Fewer works evaluated DA-STDP in an embodiment for reward maximization -- a recent survey encompassing this topic is available in \cite{bing2018survey}.


The closest previous work related to this paper are \cite{kaiser2016towards,bing2018end} and \cite{dauce2009model}.
In \cite{kaiser2016towards}, a neurorobotic lane following task is presented, where a simulated vehicle is controlled end-to-end from event-based vision to motor command.
The task is solved with an hard-coded spiking network of 16 neurons implementing a simple Braitenberg vehicle.
The performance is evaluated with respect to distance and orientation differences to the middle of the lane.
In this paper, these performance metrics are combined into a reward signal which the spiking network maximizes with the \ac{SPORE} learning rule.

In \cite{bing2018end}, the authors evaluate DA-STDP (referred to as R-STDP for reward-modulated STDP) in a similar lane following environment.
Their approach outperforms the hard-coded Braitenberg vehicle presented in \cite{kaiser2016towards}.
The two motor neurons controlling the steering receive different (mirrored) reward signals whether the vehicle is on the left or on the right of the lane.
This way, the reward provides the information of what motor command should be taken, similar to a supervised learning setup.
Conversely, the approach presented in this paper is more generic since a global reward is distributed to all synapses and does not indicate which action the agent should take.

A similar plasticity rule implenting a policy-gradient approach is derived in \cite{dauce2009model}.
Also relying on eligibility traces, this reward-learning rule uses a ``slow'' noise term to drive the exploration.
This rule is demonstrated on a target reaching task comparable to the one discussed in \cref{sec:reaching} and achieves impressive learning times (in the order of 100s) with proper tuning of the noise term.

In \cite{nakano2015spiking}, a spiking version of the free-energy-based reinforcement learning framework proposed in \cite{otsuka2010free} is introduced.
In this framework, a spiking \ac{RBM} is trained with a reward-modulated plasticity rule which decreases the free-energy of rewarding state-action pairs.
The approach is evaluated on discrete-actions tasks where the observations consist of MNIST digits processed by a pre-trained feature extractor.
However, some characteristics of \ac{RBM} are biologically implausible and make their implementation cumbersome: symmetric synapses and clocked network activity.
With our approach, network activity does not have to be manually synchronized into observation and action phases of arbitrary duration for learning to take place.

In \cite{gilra2017predicting}, a supervised synaptic learning rule named Feedback-based Online Local Learning Of Weights (FOLLOW) is introduced.
This rule is used to learn the inverse dynamics of a two-link arm -- the model predicts control commands (torques) for a given arm trajectory.
The loop is closed in \cite{gilra2017nonlinear} by feeding the predicted torques as control commands.
In contrast, \ac{SPORE} learns from a reward signal and can solve a variety of tasks.



\section{Method}
\label{sec:method}

In this section, we give a brief overview of the reward-based learning rule \ac{SPORE}.
We then discuss how \ac{SPORE} was embodied in closed-loop, along with our modifications to increase the robustness of the learned policy.

\subsection{Synaptic Plasticity with Online Reinforcement Learning (\ac{SPORE})}
\label{sec:spore}

Throughout our experiments we use an implementation of the reward-based online learning rule for spiking neural networks, named \textit{synaptic sampling}, that was introduced in \cite{Kappel2018}. The learning rule employs synaptic updates that are modulated by a global reward signal to maximize the expected reward. More precisely, the learning rule does not converge to a local maximum $\bth^*$ of the synaptic parameter vector $\bth$, but it continuously samples different solutions $\bth \sim p^*(\bth)$ from a target distribution that peaks at parameter vectors that likely yield high reward. A temperature parameter $T$ allows to make the distribution $p^*(\bth)$ flatter (high exploration) or more peaked (high exploitation).

\ac{SPORE} (\cite{spore}) is an implementation of the reward-based synaptic sampling rule \cite{Kappel2018}, that uses the NEST neural simulator (\cite{nest}). \ac{SPORE} is optimized for closed-loop applications to form an online policy-gradient approach. We briefly review here the main features of the synaptic sampling algorithm.

We consider the goal of reinforcement learning to maximize the expected future discounted reward $\mathcal{V}(\bth)$ given by
\begin{equation}
\mathcal{V}(\bth) \;=\; \expect[p(\rseq | \bth)]{ \int_{0}^\infty e^{-\frac{\tau}{\tau_e}} \,\rtau \; d \tau } \;,
\label{eqn:reward-prob-factorized}
\end{equation}
where $\rtau$ denotes the reward at time $\tau$ and $\tau_e$ is a time constant that discounts remote rewards. We consider non-negative reward $\rtau \geq 0$ at any time such that $\mathcal{V}(\bth) \geq 0$ for all $\bth$. The distribution $p(\rseq | \bth)$ denotes the probability of observing the sequence of reward $\rseq$ under a given parameter vector $\bth$. Note that computing this expectation involves averaging over a number of experimental trials and network responses.

As proposed in \cite{Kappel2018} we replace the standard goal of reinforcement learning to maximize the objective function in \cref{eqn:reward-prob-factorized} by a probabilistic framework that generates samples from the parameter vector $\bth$ according to some target distribution $\bth \sim p^*(\bth)$.
We will focus on sampling from the target distribution $p^{*}(\bth)$ of the form
\begin{equation}
p^*(\bth) \;\propto\; \ps{\bth} \, \times \, \mathcal{V}(\bth)  \; ,
\label{eq:bayes}
\end{equation}
where $\ps{\bth}$ is a prior distribution over the network parameters that allows us, for example, to introduce constraints on the sparsity of the network parameters.
It has been shown in \cite{Kappel2018} that the learning goal in \cref{eq:bayes} is achieved, if all synaptic parameters $\theta_i$ obey the stochastic differential equation
\begin{equation}
d \theta_i \;=\; \beta \, \left( \ddthetai \,\log \ps{\bth}  \, + \, \ddthetai \log \mathcal{V}(\bth)  \right)  dt  \;+ \; \sqrt{2 \beta T} \, d \wiener_{i} \;,
\label{eq:sde}
\end{equation}
where $\beta$ is a scaling parameter that functions as a learning rate, $d \wiener_{i}$ are the stochastic increments and decrements of a Wiener process and $T$ is the temperature parameter.
$\ddthetai$ denotes the partial derivative with respect to the synaptic parameter $\theta_{i}$.
The stochastic process in \cref{eq:sde} generates samples of $\bth$ that are with high probability close to the local optima of the target distribution $p^*(\bth)$.

It has been further shown in \cite{Kappel2018} that \cref{eq:sde} can be implemented using a synapse model with local update rules. The state of each synapse $i$ consists of the dynamic variables $y_i(t)$, $e_i(t)$, $g_i(t)$, $\theta_i(t)$ and $w_i(t)$.
The variable $y_i(t)$ is the pre-synaptic spike train filtered with a postsynaptic-potential kernel. 
$e_i(t)$ is the eligibility trace that maintains a brief history of pre-/post neural activity.
$g_i(t)$ is a variable to estimate the reward gradient, i.e. the gradient of the objective function in \cref{eqn:reward-prob-factorized} with respect to the synaptic parameter $\theta_i(t)$. $w_i(t)$ denotes the weight of synapse $i$ at time $t$.
In addition each synapse has access to the global reward signal $r(t)$.
The variables $e_i(t)$, $g_i(t)$ and $\theta_i(t)$ are updated by solving the differential equations:
\begin{align}
\frac{d e_i(t)}{dt} \;&=\; -\frac{1}{\tau_e} e_i(t) \,+\, w_i(t)\,y_i(t)\,(z_{post_i}(t) - \rho_{post_i}(t))  \label[equation]{eqn:synapse-sdes-e} \\
\frac{d g_i(t)}{dt} \;&=\; -\frac{1}{\tau_g} g_i(t) \,+\, r(t)\,e_i(t) \label[equation]{eqn:reward-gradient} \\ 
d \theta_i(t) \;&=\; \beta\,\bigg( c_p (\mu - \theta_i(t)) + c_g\,g_i(t) \bigg) dt \,+\, \sqrt{ 2 T_\theta \beta } \, \mathcal{W}_{i} \;, \label[equation]{eqn:synapse-sdes-th}
\end{align}
where $z_{post_i}(t)$ is a sum of Dirac delta pulses placed at the firing times of the post-synaptic neuron, $\mu$ is the prior mean of synaptic parameters ($\ps{\bth}$ in Eq.~\eqref{eq:bayes}) and $\rho_{post_i}(t)$ is the instantaneous firing rate of the post-synaptic neuron at time $t$.
The constants $c_p$ and $c_g$ are tuning parameters of the algorithm that scale the influence of the prior distribution $\ps{\bth}$ against the influence of the reward-modulated term.
Setting $c_p=0$ corresponds to a non-informative (flat) prior. In
general, the prior distribution is modeled as a Gaussian centered around $\mu$: $\ps{\bth}=\normal{\mu,\frac{1}{c_p}}\,$. We used $\mu=0$ in our simulations.
The variance of the reward gradient estimation (\cref{eqn:reward-gradient}) could be reduced by subtracting a baseline to the reward as introduced in \cite{williams1992simple}, although this was not investigated in this paper.

Finally the synaptic weights are given by the projection
\begin{equation}
w_i(t) \;=\; \begin{cases}
               w_0 \, \exp ( \theta_i(t) - \theta_0 ) & \qquad \text{if }\theta_i(t)>0 \\
               0 & \qquad \text{otherwise}
             \end{cases}\;,
\label{eqn:synapse-weight-map}
\end{equation}
which scaling and offset parameters $w_0$ and $\theta_0$, respectively.

In \ac{SPORE} the differential equations \crefrange{eqn:synapse-sdes-e}{eqn:synapse-sdes-th} are solved using the Euler method with a time step of 1~ms.
The dynamics of the postsynaptic term $y_i(t)$, the eligibility trace $e_i(t)$ and the reward gradient $g_i(t)$ are updated at each time step.
The dynamics of $\theta_i(t)$ and $w_i(t)$ are updated on a coarser time grid with step width 100~ms for the sake of simulation speed.
The synaptic weights remain constant between two updates.
Synaptic parameters are clipped at $\theta_{min}$ and $\theta_{max}$.
Parameter gradients $g_i(t)$ are clipped at $\pm \Delta\theta_{max}$.
The parameters used in our evaluation are stated in \crefrange{table:nest}{table:music}.

\subsection{Closed-Loop Embodiment Implementation}

\begin{figure}
  \centering
  \includegraphics[width=0.49\textwidth]{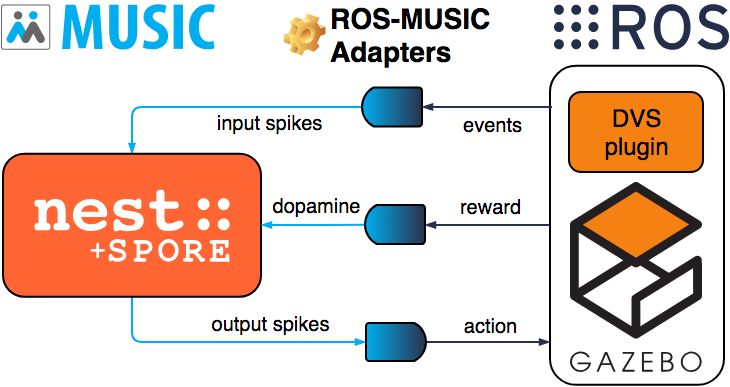}~
  \includegraphics[width=0.49\textwidth]{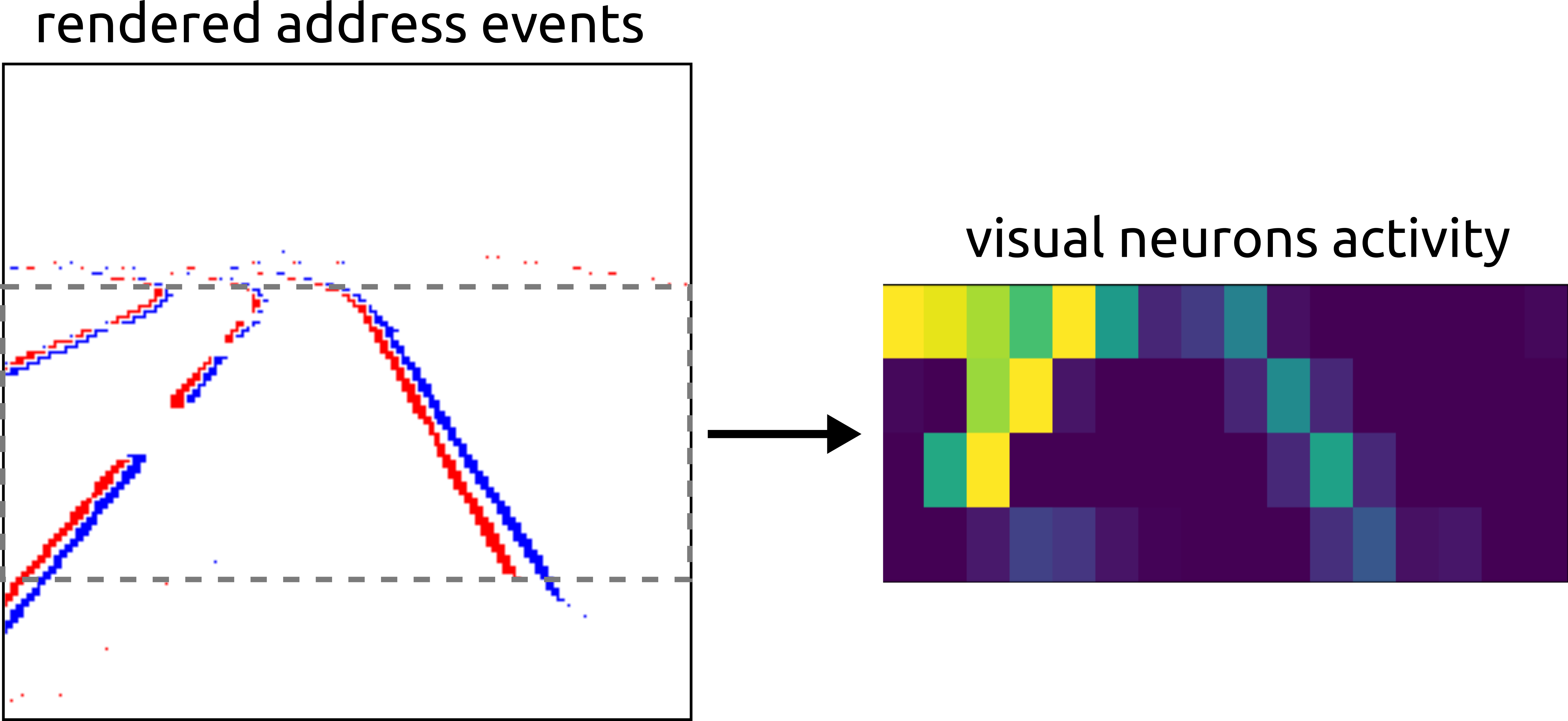}
  \caption{Implementation of the embodied closed-loop evaluation of the reward-based learning rule \ac{SPORE}.
    Left: our asynchronous framework based on open-source software components.
    The spiking network is implemented with the NEST neural simulator (\cite{nest}), which communicates spikes with MUSIC (\cite{Djurfeldt2010,Ekeberg2008}).
    The reward is streamed to all synapses in the spiking network learning with SPORE (\cite{spore}).
    Spikes are encoded from address events and decoded to motor commands with ROS-MUSIC tool-chain adapters (\cite{Weidel2016}).
    Address events are emitted by the DVS plugin (\cite{kaiser2016towards}) within the simulated robotic environment Gazebo (\cite{gazebo}), which communicates with ROS (\cite{ros}).
    Right: Encoding visual information to spikes for the lane following experiment, see \cref{sec:lane-follow} for more information.
    Address events (red and blue pixels on the rendered image) are downscaled and fed to visual neurons as spikes.
  }
  \label{fig:implementation}
\end{figure}

Usually, synaptic learning rules are solely evaluated on open-loop pattern classification tasks \cite{zenke2018superspike,Neftci17_stocsyna,Pfister_etal06_optispik,Urbanczik_Senn14_learby}.
An embodied evaluation is technically more involved and requires a closed-loop environment simulation.
A core contribution of this paper is the implementation of a framework allowing to evaluate the validity of bio-plausibe plasticity models in closed-loop robotics environments.
We rely on this framework to evaluate the synaptic sampling rule \ac{SPORE} (\cite{spore}), as depicted in \cref{fig:implementation}.
n
This framework is tailored for evaluating spiking network learning rules in an embodiment.
Visual sensory input is sensed, encoded as spikes, processed by the network, and output spikes are converted to motor commands.
The motor commands are executed by the agent, which modifies the environment.
This modification of the environment is sensed by the agent.
Additionally, a continuous reward signal is emitted from the environment.
\ac{SPORE} tries to maximize this reward signal online by steering the ongoing synaptic plasticity processes of the network towards configurations which are expected to yield more overall reward.
Unlike classical reinforcement learning setup, the spiking network is treated as a dynamical system continuously receiving input and outputting motor commands.
This allows us to report learning progress with respect to (biological) simulated time, unlike classical reinforcement learning which reports learning progress in number of iterations.
Similarly, we reset the agent only when the task is completed (in the reaching task) or when the agent goes off-track (in the lane following task).
We do not enforce finite-time episodes and neither the agent nor \ac{SPORE} are notified of the reset.

This framework relies on many open-source software components:
As neural simulator we use NEST (\cite{nest}) combined with the open-source implementation of \ac{SPORE} (\cite{Kappel2018}\footnote{\url{https://github.com/IGITUGraz/spore-nest-module}}).
The robotic simulation is managed by Gazebo (\cite{gazebo}) and ROS (\cite{ros})
and visual perception is realized using the open-source DVS plugin for Gazebo (\cite{kaiser2016towards}\footnote{\url{https://github.com/HBPNeurorobotics/gazebo_dvs_plugin}}).
This plugin emits polarized address events when variations in pixel intensity cross a threshold.
The robotic simulator and the neural network run in different processes.
We rely on MUSIC (\cite{Djurfeldt2010,Ekeberg2008}) to communicate and transform the spikes and we employ the ROS-MUSIC tool-chain by \cite{Weidel2016} to bridge between the two communication frameworks.
The latter also synchronizes ROS time with spiking network time.
Most of these components are also integrated in the Neurorobotics Platform (NRP) \cite{falotico2017connecting}, except for MUSIC and the ROS-MUSIC tool-chain.
Therefore, the NRP does not support streaming a reward signal to all synapses, required in our experiments.

As part of this work, we contributed to the Gazebo DVS plugin by integrating it to ROS-MUSIC, and to the \ac{SPORE} module by integrating it with MUSIC.
These contributions enable researchers to design new ROS-MUSIC experiments using event-based vision to evaluate \ac{SPORE} or their own biologically-plausible learning rules.
A clear advantage of this framework is that the robotic simulation can be substituted for a real robot seamlessly.
However, the necessary human supervision in real robotics coupled with the many hours needed by \ac{SPORE} to learn a performing policy is currently prohibitive.
The simulation of the whole framework was conducted on a Quad core Intel Core i7-4790K with 16GB RAM in real-time.

\subsection{Learning Rate Annealing}
\label{sec:annealing}
In the original work presenting \ac{SPORE} (\cite{Kappel2018,Kappel2015,Kappel2014,Yu2016a}), the learning rate $\beta$ and the temperature $T$ were kept constant throughout the learning process.
Note that in deep learning, learning rates are often regulated by the optimization processes (\cite{Kingma_Ba14_adammeth}).
We found that the learning rate $\beta$ of \ac{SPORE} plays an important role in learning and benefit from an annealing mechanism.
This regulation allows the synaptic weights to converge to a stable configuration and prevents the network to forget previous policy improvements.
For the lane following experiment presented in this paper, the learning rate $\beta$ is decreased over time, which also reduces the temperature (random exploration), see \cref{eq:sde}.
Specifically, we decay the learning rate $\beta$ exponentially with respect to time:
\begin{equation}
  \frac{d \beta(t)}{dt} = - \lambda \beta(t).
\end{equation}
\noindent The learning rate is updated following this equation every 10 minutes.
Independently decaying the temperature term $T$ was not investigated, however we expect a minor impact on the performance because of the high variance of the reward gradient estimation, intrinsically leading the agent to explore.

\section{Evaluation}\label{sec:eval}

We evaluate our approach on two neurorobotic tasks: a reaching task and the lane following task presented in \cite{kaiser2016towards,bing2018end}.
In the following sections, we describe these tasks and the ability of \ac{SPORE} to solve them.
Additionally, we analyze the performance and stability of the learned policies with respect to the prior distribution $\ps{\bth}$ and learning rate $\beta$, see \cref{eq:sde}.

\subsection{Experimental Setup}

\begin{figure}
    \centering
    \includegraphics[width=.48\textwidth,trim={0 0 0 9cm},clip=true]{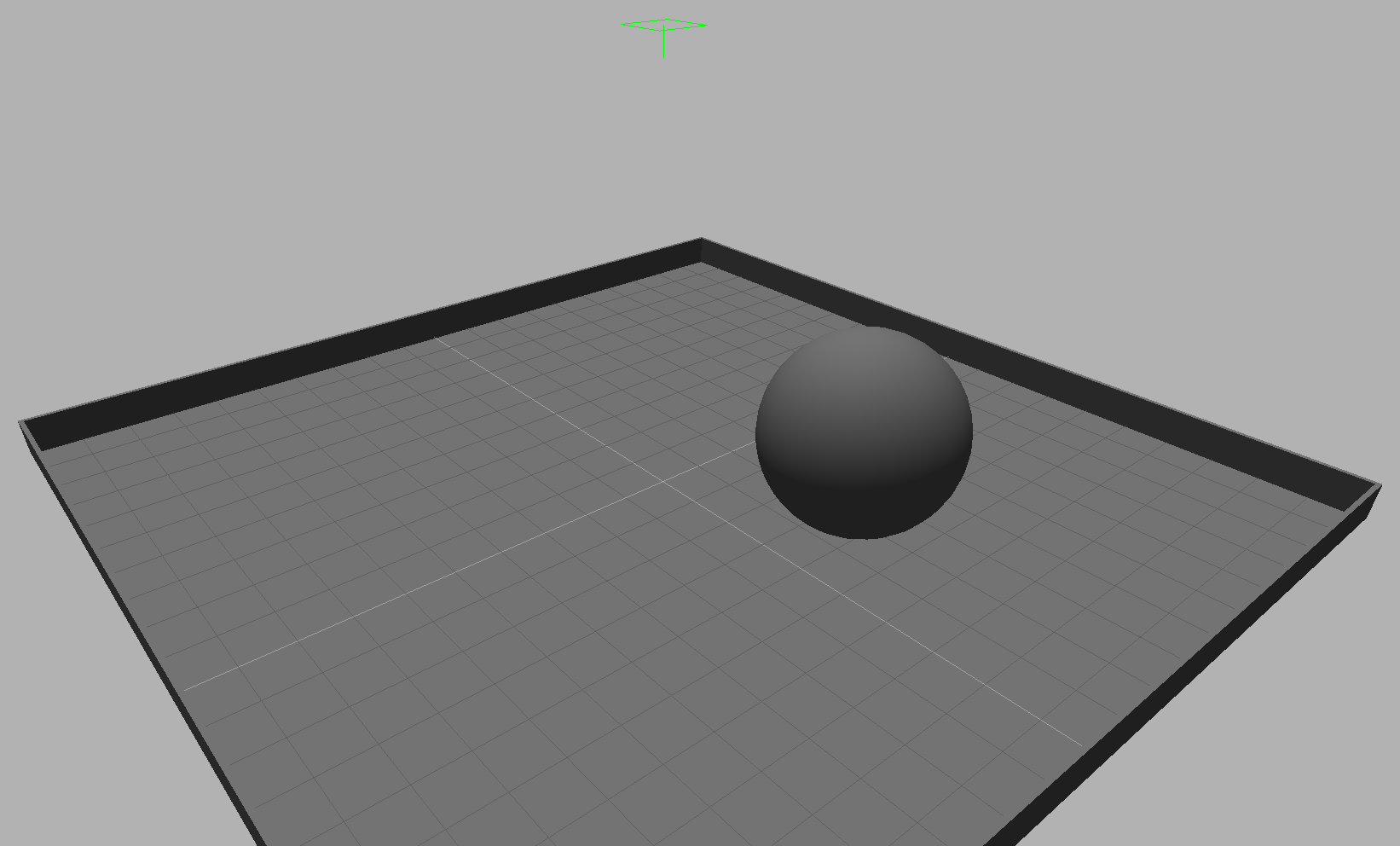}
    \includegraphics[width=.48\textwidth]{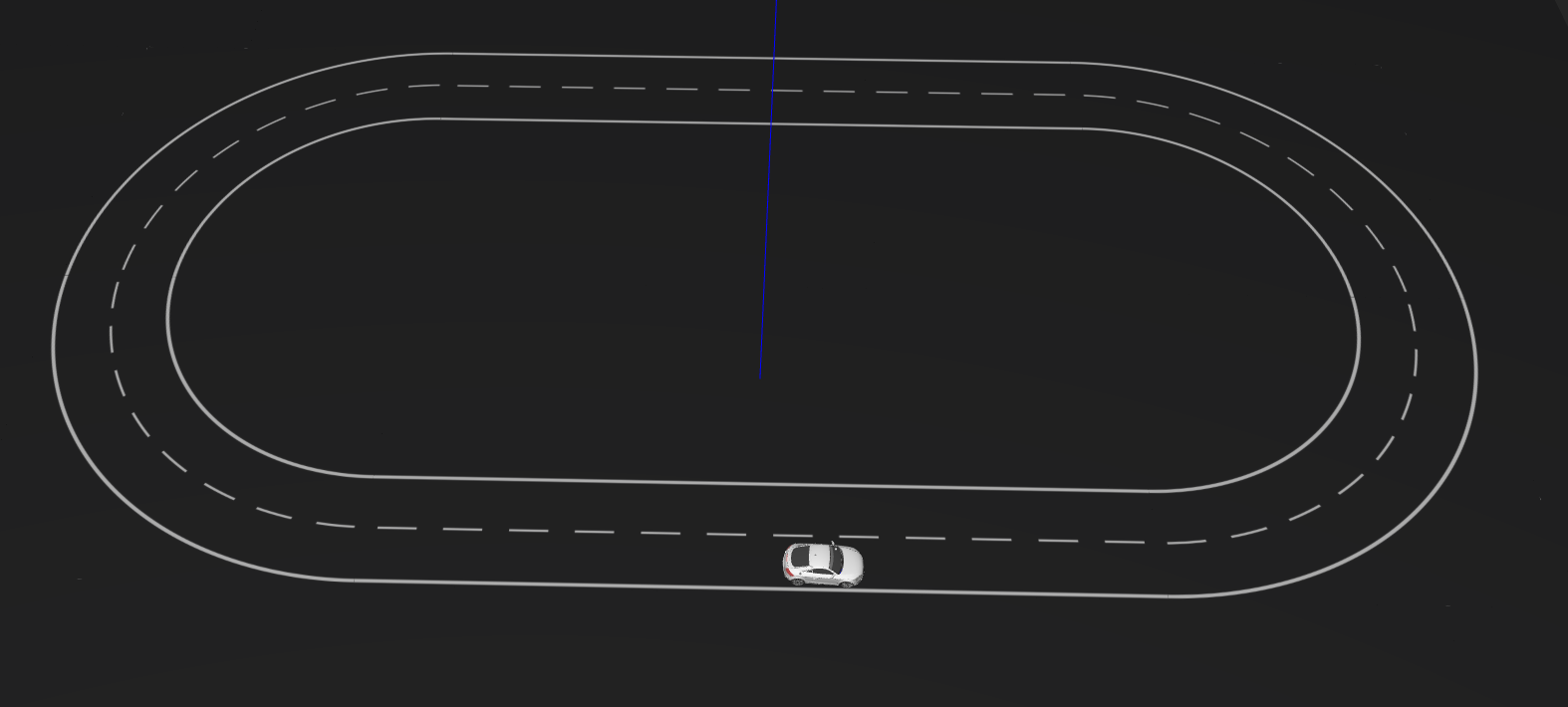}
    \caption{
      Visualization of the setup for the two experiments.
      Left: reaching experiment.
      The goal of the task is to control the ball to the center of the plane.
      Visual input is provided by a DVS simulation above the plane looking downward.
      The ball is controlled with Cartesian velocity vectors.
      Right: Lane following experiment.
      The goal of the task is to keep the vehicle on the right lane of the road.
      Visual input is provided by a DVS simulation attached to the vehicle looking forward to the road.
      The vehicle is controlled with steering angles.
    }
    \label{fig:setup}
\end{figure}

The tasks used for our evaluation are depicted in \cref{fig:setup}.
In both tasks, a feed-forward all-to-all two-layers network of spiking neurons is trained with \ac{SPORE} to maximize a task-specific reward.
Previous work has shown that this architecture was sufficient for the task complexity considered \cite{kaiser2016towards,bing2018end,dauce2009model}.
The network is end-to-end and maps the address events of a simulated DVS to motor commands.
The parameters used for the evaluation are presented in \crefrange{table:nest}{table:music}.
In the next paragraphs, we describe the tasks together with their decoding schemes and reward functions.


\subsubsection{Reaching Task}\label{sec:reaching}

The reaching task is a natural extension of the open-loop blind reaching task on which \ac{SPORE} was evaluated in \cite{Yu2016a}.
A similar visual tracking task was presented in \cite{dauce2009model}, with a different visual input encoding.
In our setup, the agent controls a ball of 2m radius which has to move towards the 2m radius center of a 20mx20m plane enclosed with walls.
Sensory input is provided by a simulated DVS with a resolution of 16x16 pixels located above the center which perceives the ball and the entire plane.
There is one visual neuron corresponding to each DVS pixel -- we make no distinctions between ON and OFF events.
We additionally enhance the input space with an axis feature neuron for each row and each column.
These neurons fire for each spikes in the respective row or column of neurons they cover.
Both 16x16 visual neurons and 2x16 axis feature neurons are connected to all 8 motor neurons with 10 plastic \ac{SPORE} synapses, resulting in 23040 learnable parameters.
The network controls the ball with instantaneous velocity vectors through the Gazebo Planar Move Plugin.
Velocity vectors are decoded from output spikes with the linear decoder:
\begin{equation}
  \begin{split}
    v &=
    \begin{bmatrix}
      \dot{x}\\
      \dot{y}\\
    \end{bmatrix}
    =
    \begin{bmatrix}
      cos(\beta_1) & cos(\beta_2) & \hdots & cos(\beta_N)\\
      sin(\beta_1) & sin(\beta_2) & \hdots & sin(\beta_N)
    \end{bmatrix}
    \begin{bmatrix}
      a_1\\
      a_2\\
      \vdots\\
      a_N
    \end{bmatrix}\\
    \beta_k &= \frac{2k\pi}{N} ,
    \end{split}
\end{equation}
\noindent with $a_k$ the activity of motor neuron $k$ obtained by applying a low-pass filter on the spikes with time constant $\tau$.
This decoding scheme consists of equally distributing $N$ motor neurons on a circle representing their contribution to the displacement vector.
For our experiment, we set $N=8$ motor neurons.
We add an additional exploration neuron to the network which excites the motor neurons and is inhibited by the visual neurons.
This neuron prevents long periods of immobility.
Indeed, when the agent decides to stay motionless, it does not receive any sensory input as the DVS simulation only senses change.
Since the network is feedforward, the absence of sensory input causes the neural activity to drop, leading to more immobility.

The ball is reset to a random position on the plane if it has reached the center.
This reset is not signaled to the network -- aside from the abrupt change in visual input -- and does not mark the end of an episode.
Let $\beta_\text{err}$ denote the absolute value of the angle between the straight line to the goal and the direction taken by the ball.
The agent is rewarded if the ball moves in the direction towards the goal $\beta_\text{err} < \beta_\text{lim}$ at a sufficient velocity $v > v_\text{lim}$.
Specifically, the reward $r(t)$ is computed as:
\begin{equation}
  \begin{split}
    r(t) &= 35  \sqrt{r_v} (r_\beta + 1)^5\\
    r_\beta &=
    \begin{cases}
      1 - \frac{\beta_\text{err}}{\beta_\text{lim}}, & \text{if}\ \beta_\text{err} < \beta_\text{lim} \\
      0, & \text{otherwise}
    \end{cases}\\
    r_v &=
    \begin{cases}
      |v|, & \text{if}\ |v| > v_\text{lim}\\
      0, & \text{otherwise}
    \end{cases}
    .
  \end{split}
\end{equation}
\noindent This signal is smoothed with an exponential filter before being streamed to the agent.
This formulation provides a continuous feedback to the agent, unlike delivering a discrete terminal reward upon reaching the goal state.
In our experiments, discrete terminal rewards did not suffice for the agent to learn performing policies in a reasonable amount of time.
On the other hand, distal rewards are supported by \ac{SPORE} through eligibility traces, as was demonstrated in \cite{Kappel2018,Yu2016a} for open-loop tasks with clearly delimited episodes.
This suggests that additional mechanisms or hyperparameter tuning would be required for \ac{SPORE} to learn from distal rewards online.



\subsubsection{Lane following Task}\label{sec:lane-follow}

The lane following task was already used to demonstrate spiking neural controllers in \cite{kaiser2016towards} and \cite{bing2018end}.
The goal of the task is to steer a vehicle to stay on the right lane of a track.
Sensory input is provided by a simulated DVS with a resolution of 128x32 pixels mounted on top of the vehicle showing the track in front.
There are 16x4 visual neurons covering the pixels, each neuron responsible for a 8x8 pixel window.
Each visual neuron spikes at a rate correlated to the amount of events in its window, see \cref{fig:implementation}.
The vehicle starts driving on a fixed starting point with a constant velocity on the right lane of the track.
As soon as the vehicle leaves the track, it is reset to the starting point.
As in the reaching task, this reset is not explicitly signaled to the network and does not mark the end of a learning episode.

The network controls the angle of the vehicle by steering it, while its linear velocity is constant.
The output layer is separated into two neural populations.
The steering commands sent to the agent consist of the difference of activity between these two populations.
Specifically, steering commands are decoded from output spikes as a ratio between the following linear decoders:
\begin{equation}
  \begin{split}
  a_L &= \sum_{i=1}^{N/2} a_i, \\
  a_R &= \sum_{i=N/2}^{N} a_i, \\
  r &= \frac{a_L - a_R}{a_L + a_R}.
  \end{split}
\end{equation}
\noindent The first $N/2$ neurons pull the steering on one side, while the remaining $N/2$ neurons pull steering to the other side.
We set $N=8$ so that there are $4$ left motor neurons and $4$ right motor neurons.
The steering command is obtained by discretizing the ratio $r$ into five possible commands: hard left (-30\degree), left (-15\degree), straight (0\degree), right (15\degree) and hard right (30\degree).
The decision boundaries between these steering angles are $r=\{-10, -2.5, 2.5, 10\}$ respectively.
This discretization is similar than the one used in \cite{Wolf2017}.
It yielded better performance than directly using $r$ (multiplied with a scaling constant $k$) as a continuous-space steering command as in \cite{kaiser2016towards}.

The reward signal delivered to the vehicle is equivalent to the performance metrics used in \cite{kaiser2016towards} to evaluate the policy.
As in the reaching task, the reward depends on two terms -- the angular error $\beta_\text{err}$ and the distance error $d_\text{err}$.
The angular error $\beta_\text{err}$ is the absolute value of the angle between the right lane and the vehicle.
The distance error $d_\text{err}$ is the distance between the vehicle and the center of the right lane.
The reward $r(t)$ is computed as:
\begin{equation}
  r(t) = e^{\displaystyle -0.03~\beta_\text{err}^2} \times e^{\displaystyle -70~d_\text{err}^2}.
\end{equation}
\noindent The constants are chosen so that the score is halved every 0.1m distance error or 5\degree angular error.
Note that this reward function is comprised between $[0, 1]$ and is less informative than the error used in \cite{bing2018end}.
In our case, the same reward is delivered to all synapses, and a particular reward value does not indicate whether the vehicle is on the left or on the right of the lane.
The decay of the learning rate is $\lambda=\num{8.5e-5}$, see \cref{table:spore}.

\subsection{Results}

Our results show that \ac{SPORE} is capable of learning policies online for moderately difficult embodied tasks within some simulated hours.
We first discuss the results on the reaching task, where we evaluated the impact of the prior distribution.
We then present the results on the lane following task, where the impact of the learning rate was evaluated.

\subsubsection{Impact of Prior Distribution}
\begin{figure}
    \centering
    \includegraphics[width=0.48\textwidth]{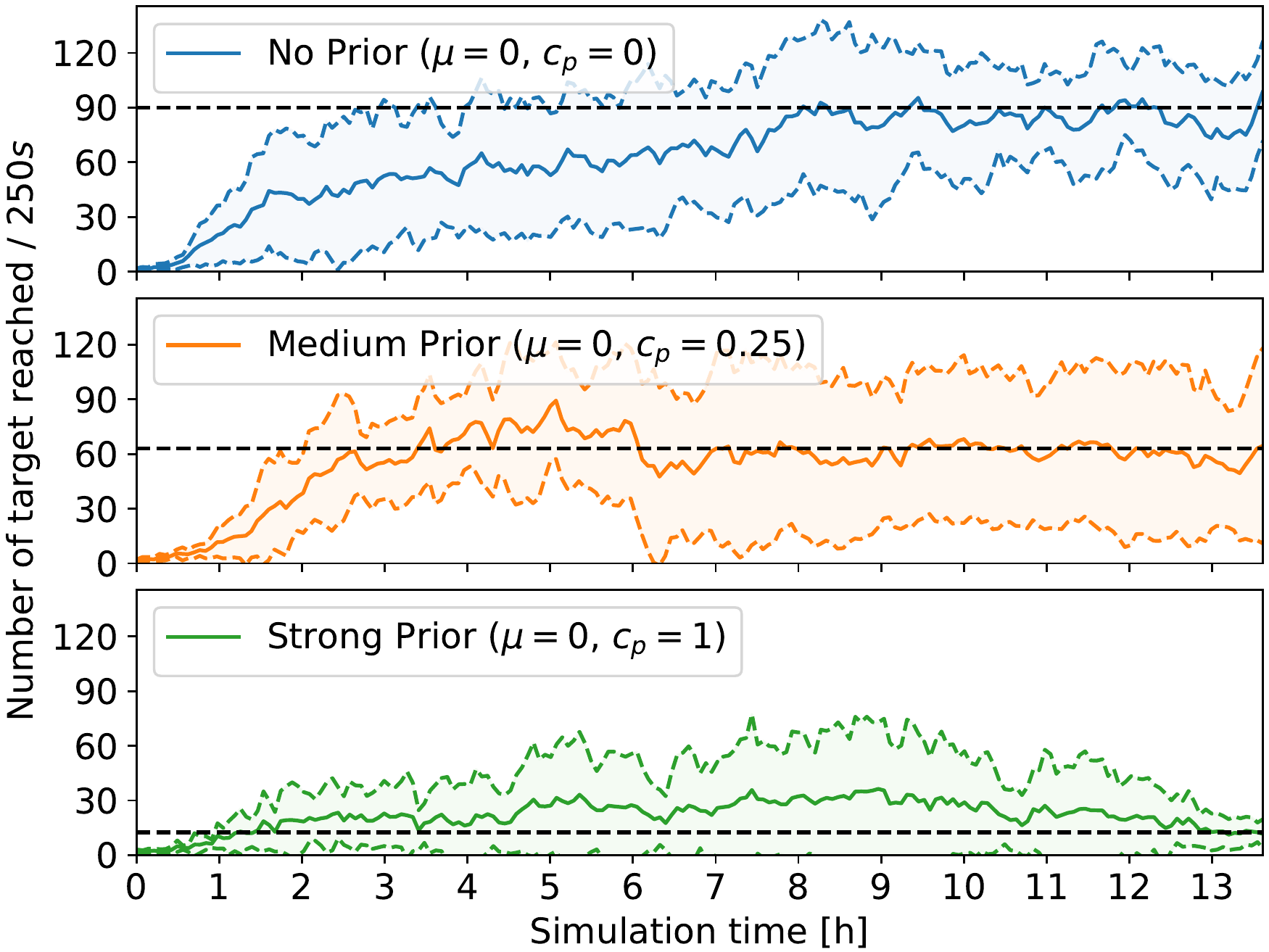}
    \includegraphics[width=0.48\textwidth]{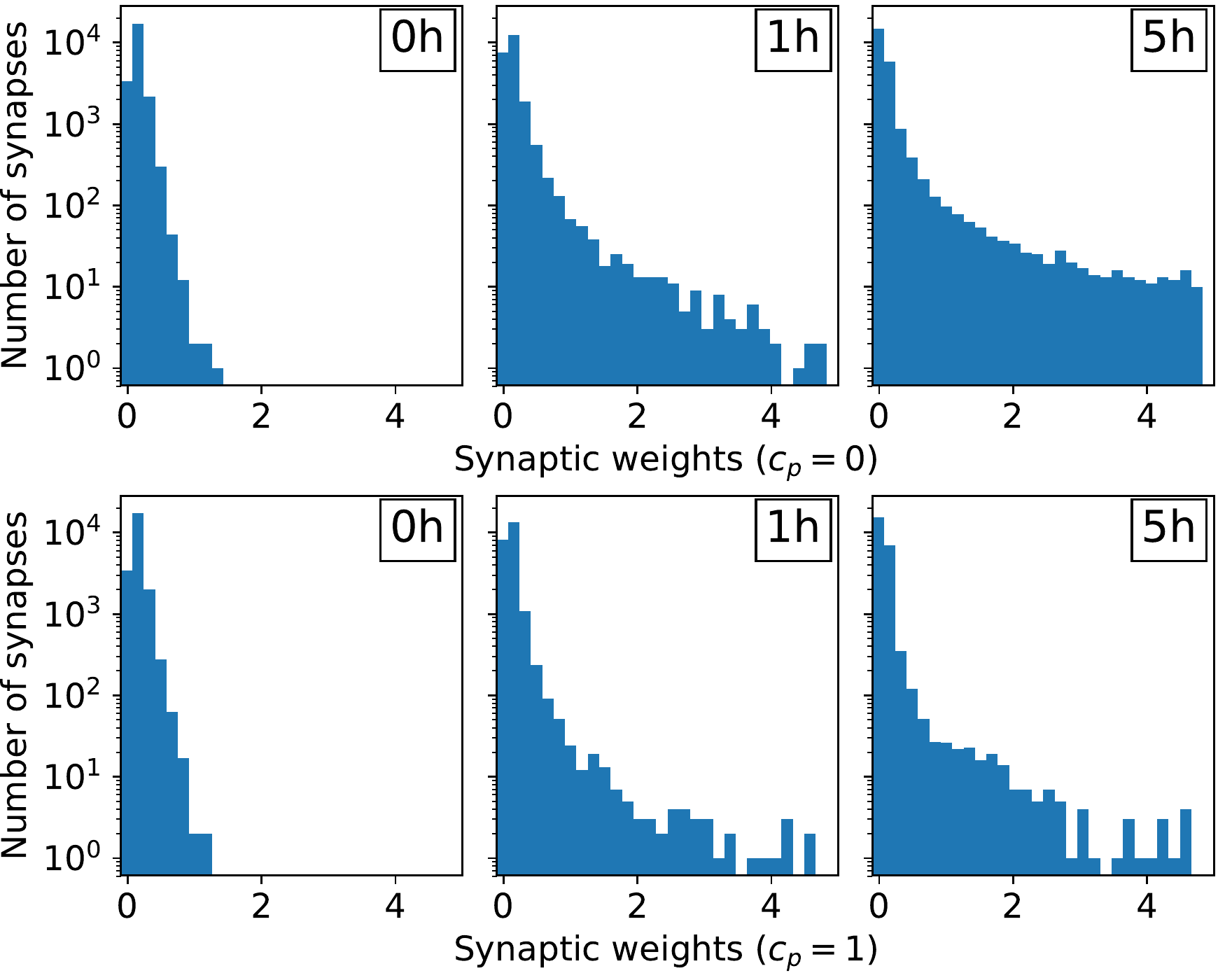}
    \caption{Results for the reaching task.
      Left: comparing the effect of different prior configurations on the overall learning performance.
      The results were averaged over 8 trials.
      The performance is measured with the rate at which the target is reached (the ball moves to the center and is reset at a random position).
      Right: Development of the synaptic weights over the course of learning for two trials: no prior ($c_p = 0$, top) and strong prior ($c_p=1$, bottom).
      In both cases, the number of weak synaptic weights (below 0.07) increases significantly over time.
    }
    \label{fig:results_target_reaching}
\end{figure}

For the reaching task, a flat prior $c_p=0$ yielded the policy with highest performance, see \cref{fig:results_target_reaching}.
In this case, the performance improves rapidly within a few hours of simulated time, and the ball reaches the center about 90 times every \SI{250}{\second}.
Conversely, a strong prior ($c_p=1$) forcing the synaptic weights close to $0$ prevented performing policies to emerge.
In this case, after 13h of learning, the ball reaches the center only about 10 times on average every 250s, a performance comparable to the random policy.
Less constraining priors also affected the performance of the learned policies compared to the unconstrained case, but allowed learning to happen.
With $c_p=0.25$, the ball reaches the center about 60 times on average every \SI{250}{\second}.
Additionally, the number of retracting synapses increases over time -- even in the flat prior case -- reducing the computational overhead, important for a neuromorphic hardware implementation (\cite{bellec2017deep}).
Indeed, for $c_p=0$, the number of weak synaptic weights (below $0.07$) increased from 3329 to 7557 after 1h of learning to 14753 after 5h of learning (out of 23040 synapses in total).
In other words, only 36\% of all synapses are active.
The weight distribution for $c_p=0.25$ is similar to the no-prior case $c_p=0$.
The strong prior $c_p=1$ prevented strong weights to form, trading-off performance.
The same trend is observed for the lane following task, where only 33\% of all synapses are active after 4h of learning, see \cref{fig:results_lane_following}.


\begin{figure}
  \centering
  \includegraphics[width=0.8\textwidth]{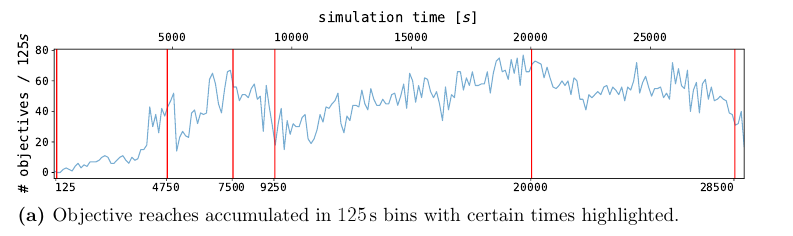}
  \includegraphics[width=0.8\textwidth]{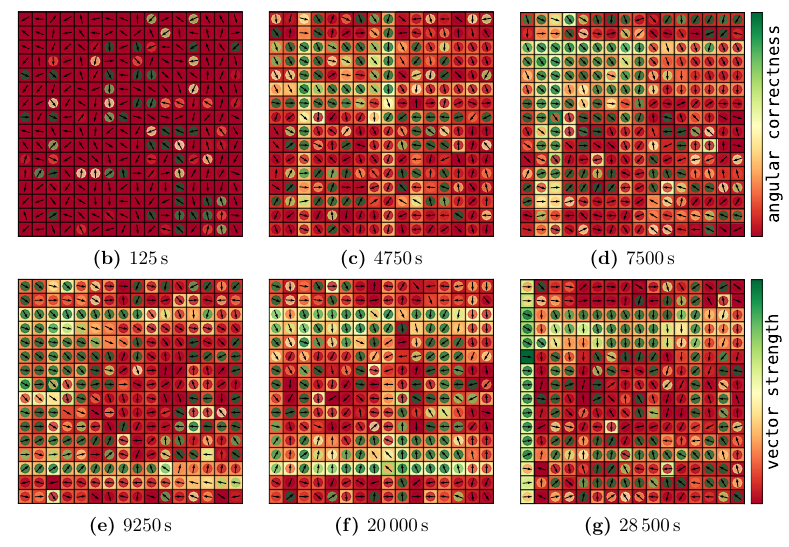}
  \caption[Reaching -- policy development for single trial]{%
    Policy development for selected points in time in a single trial.
    On the top, the performance over time for a single, well-performing trial is depicted.
    The red lines indicate certain points in time, for which the policies are shown in the bottom $6$ figures.
    Each policy plot consists of a 2d-grid representing the DVS pixels.
    Hereby, every pixel contains a vector, which indicates the motion corresponding
    to the contribution of an event emitted by this pixel.
    The magnitude of the contribution (vector strength) is indicated by the outer pixel area.
    The inner circle color represents the assessment of the vector direction (angular correctness).
    \label{fig:trial_target_reaching}
  }
\end{figure}

The analysis of a single trial with $c_p=0.25$ is depicted in \cref{fig:trial_target_reaching}.
The performance does not converge and rather rise and drop while the network is sampling configurations.
On initialization (b), the policy employs weak actions with random directions.

After over \SI{4750}{\second} of learning (c), the first local maximum is reached.
Vector directions have largely turned towards the grid center (see inner pixel colors).
Additionally, the overall magnitude of the weights has largely increased, as could be expected from the weight histogram in \cref{fig:results_target_reaching}.
In particular, patterns of single rows and columns emerge, due to the 2x16 axis feature neurons described in \cref{sec:reaching}.
One drawback of the axis feature neurons can be seen in the center column of pixel.
The axis feature neuron responsible for this column learned to push the ball down, since the ball mostly visited the upper part of the grid.
However, at the center, the correct direction to push the ball towards the center is flipped.

At \SI{7500}{\second} (d), the performance has further increased.
The policy, as shown in the second peak has grown even stronger for many pixels which also point in the right direction.
The pixels pointing in the wrong direction mostly have a low vector strength.

After \SI{9250}{\second} (e), the performance drops to half its previous performance.
As we can see from the policy, the weights grew even stronger.
Some strong pixels vectors pointing towards each other have emerged, which can lead to the ball constantly moving up and down, without receiving any reward.

After this valley, the performance rises slowly again and at \SI{20000}{\second} of simulation time (f) the policy has reached the maximum performance of this trial.
Around the whole grid, strong motion vectors push the ball towards the center, and the ball reaches the center around $140$ times every \SI{250}{\second}.

Just before the end of the trial, the performance drops again (g).
Most vectors still point towards the right direction, however, the overall strength has largely decreased.

\subsubsection{Impact of Learning Rate}
\begin{figure}
    \centering
    \includegraphics[width=0.48\textwidth]{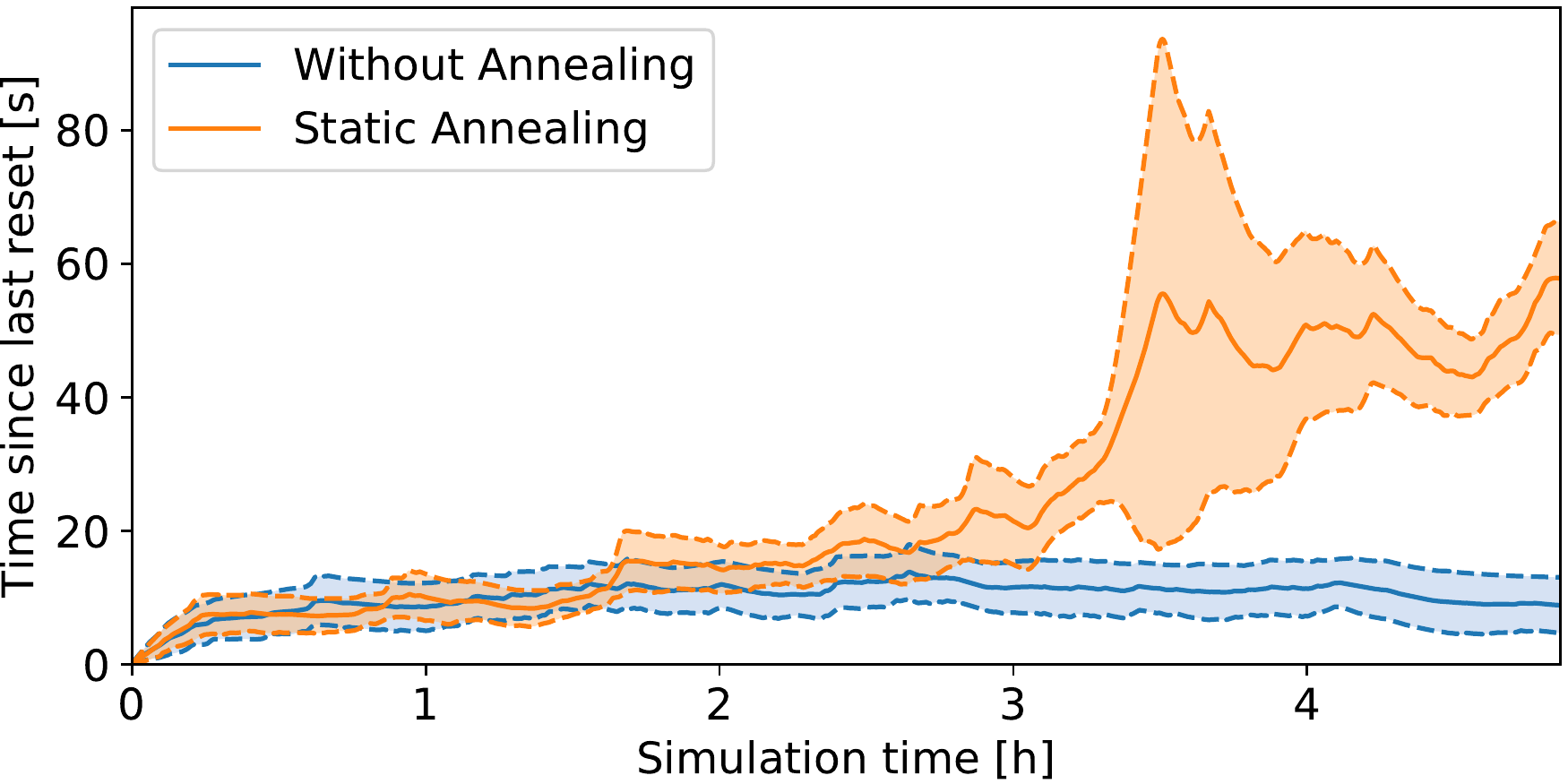}
    \includegraphics[width=0.48\textwidth]{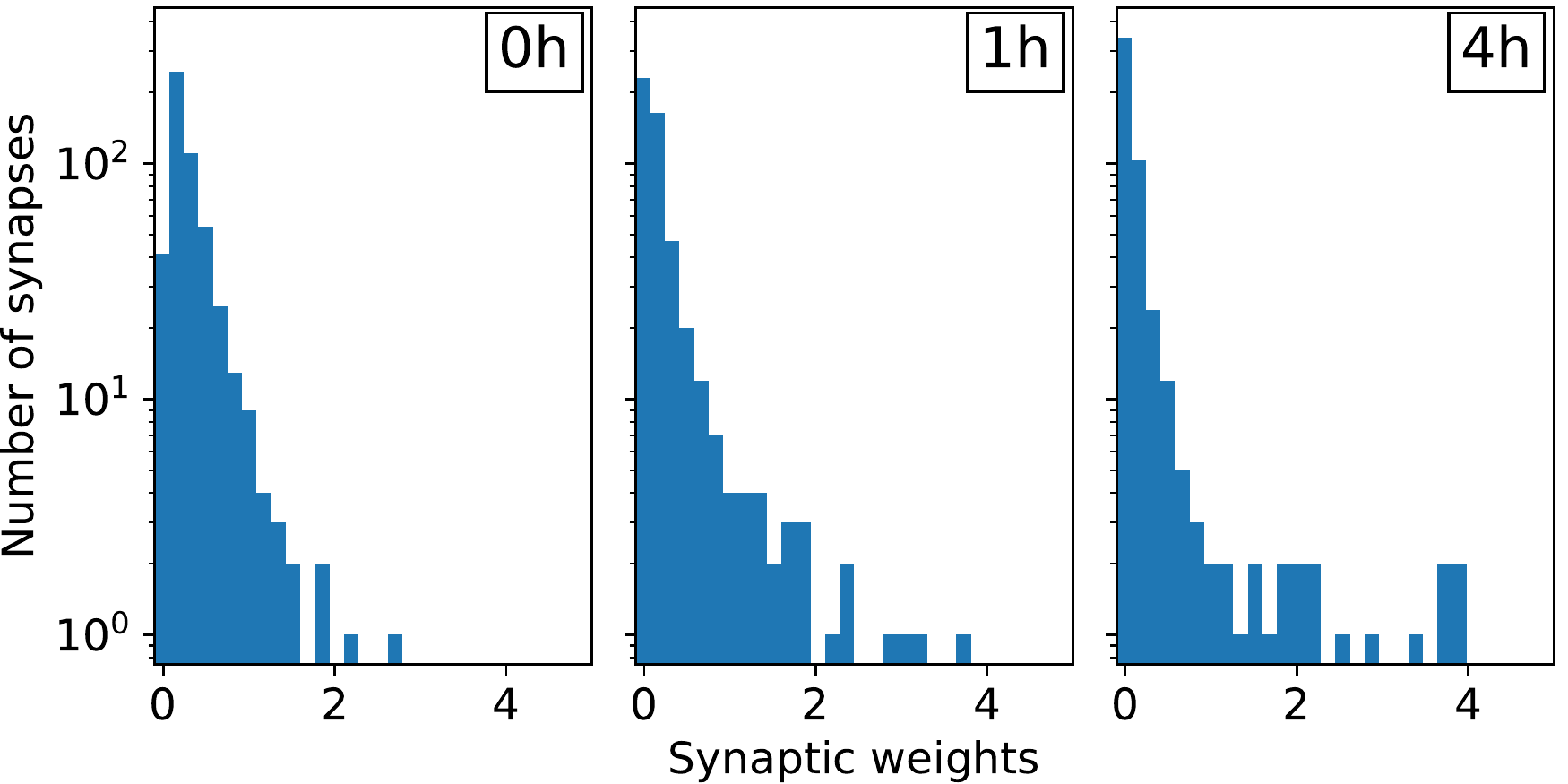}
    \caption{Results for the lane following task with a medium prior ($c_p=0.25$).
      Left: comparing the effect of annealing on the overall learning performance.
      The results were averaged over 6 trials.
      Without annealing, performance improvements are not retained and the network does not learn to perform the task.
      With annealing, the learning rate $\beta$ decreases over time and performance improvements are retained.
      Right: Development of the synaptic weights over the course of learning for a medium prior of $c_p=0.25$ with annealing.
      The number of weak synaptic weights (below $0.07$) increases from 41 to 231 after 1h of learning to 342 after 4h of learning (out of 512 synapses in total).
    }
    \label{fig:results_lane_following}
\end{figure}

For the lane following experiment, we show that the learning rate $\beta$ plays an important role for retaining policy improvements.
Specifically, when the learning rate $\beta$ remains constant over the course of learning, the policy does not improve compared to random, see \cref{fig:results_lane_following}.
In the random case, the vehicle remains about 10 seconds on the right lane until triggering a reset.
After about 3h of learning, the learning rate $\beta$ decreased to 40\% of its initial value and the policy starts to improve.
After 5h of learning, the learning rate $\beta$ approaches 20\% of its initial value and the performance improvements are retained.
Indeed, while the weights are not frozen, the amplitude of subsequent synaptic updates are drastically reduced.
In this case, the policy is significantly better than random and the vehicle remains on the right lane about 60s on average.

\section{Conclusion}\label{sec:conclusion}
The endeavor to understand the brain spans over multiple research fields.
Collaborations allowing synaptic learning rules derived by theoretical neuroscientists to be evaluated in closed-loop embodiment are an important milestone of this endeavor.
In this paper, we successfully implemented a framework allowing this evaluation by relying on open-source software components for spiking network simulation \cite{nest,spore}, synchronization and communication \cite{Djurfeldt2010,Ekeberg2008,Weidel2016,ros} and robotic simulation \cite{gazebo,kaiser2016towards}.
The resulting framework is capable of learning online the control of simulated and real robots with a spiking network in a modular fashion.
This framework is used to evaluate the reward-learning rule \ac{SPORE}~(\cite{Kappel2018,Kappel2015,Kappel2014,Yu2016a}) on two closed-loop visuomotor tasks.
Overall, we have shown that \ac{SPORE} was capable of learning shallow feedforward policies online for moderately difficult embodied tasks within some simulated hours.
This evaluation allowed us to characterize the influence of the prior distribution on the learned policy.
Specifically, constraining priors deteriorate the performance of the learned policy but prevent strong synaptic weights to emerge, see \cref{fig:results_target_reaching}.
Additionally, for the lane following experiment, we have shown how learning rate regulation enabled policy improvements to be retained.
Inspired by simulated annealing, we presented a simple method decreasing the learning rate over time.
This method does not model a particular biological mechanism, but seems to work better in practice.
On the other hand, novelty is known to modulate plasticity through a number of mechanisms (\cite{rangel2016neurotransmitters,hamid2016mesolimbic}).
Therefore, a decrease in learning rate after familiarization with the task is reasonable.

On a functional scale, deep learning methods still outperform biologically plausible learning rules such as \ac{SPORE}.
For future work, the performance gap between \ac{SPORE} and deep learning methods should be tackled by taking inspiration from deep learning methods.
Specifically, the online learning method inherent to \ac{SPORE} is impacted by the high variance of the policy evaluation.
This problem was alleviated in policy-gradient methods by introducing a critic trained to estimate the expected return of a given state.
This expected return is used as a baseline which reduces the variance of the policy evaluation.
Decreasing the variance could also be achieved by considering an action-space noise as in \cite{dauce2009model} instead of a parameter-space noise implemented by the Wiener process in \cref{eq:sde}.
Lastly, an automatic mechanism to regulate the learning rate $\beta$ is beneficial for more complex task.
Such a mechanism could be inspired by trust-region methods (\cite{schulman2015trust}), which constrains weight updates to alter the policy little by little.
These improvements should increase \ac{SPORE} performance so that more complex tasks such as multi-joint effector control and discrete terminal rewards -- supported by design by the proposed framework -- could be considered.

\section*{Conflict of Interest Statement}
The authors declare that the research was conducted in the absence of any commercial or financial relationships that could be construed as a potential conflict of interest.

\section*{Author Contributions}
All the authors participated in writing the paper.
JK, MH, AK, JCVT and DK conceived the experiments and analyzed the data.

\section*{Funding}
This research has received funding from the European Union's Horizon 2020 Framework Programme for Research and Innovation under the Specific Grant Agreement No. 720270 (Human Brain Project SGA1) and No. 785907 (Human Brain Project SGA2), as well as a fellowship within the FITweltweit programme of the German Academic Exchange Service (DAAD) [MH].
In addition, this work was supported by the H2020-FETPROACT project Plan4Act (\#732266) [DK].

\section*{Acknowledgments}
The collaboration between the different institutes that led to the results reported in the present paper was carried out under CoDesign Project 5 (CDP5 -- Biological Deep Learning) of the Human Brain Project.

\section*{Data Availability Statement}
No datasets were generated for this study.

\begin{params}{NEST Parameters}\label{table:nest}
  time-step/resolution       & $\SI{1}{\ms}$ \\
  synapse update interval   & $\SI{100}{\ms}$ \\
  (reaching) exploration noise         & $\SI{35}{\hertz}$ \\
  (reaching) noise to exploration exc. & $750.0$ \\
  (reaching) visual to exploration inh.& $\normal{-500, 50}$ \\
  (reaching) exploration to motor exc. & $10.0$ \\
\end{params}

\begin{params}{SPORE Parameters}\label{table:spore}
  visual to motor exc.      & $\normal{0.8, 0.6}$ (clipped at $0$) \\
  visual to motor mul.      & $10$ \\
  temperature ($T$)               & $0.1$ \\
  initial learning rate ($\beta$)   & $\num{1e-7}$ \\
  learning rate decay ($\lambda$)   & $\num{8.5e-5}$ \\
  integration time          & \SI{50}{\second} \\
  max synaptic parameter $(\theta_{max}$)   & $5.0$ \\
  min synaptic parameter $(\theta_{min}$)    & $-2.0$ \\
  (reaching) episode length & \SI{1}{\second} \\
  (lane following) episode length & \SI{2}{\second} \\
\end{params}

\begin{params}{ROS-MUSIC Parameters}\label{table:music}
  MUSIC time-step         & $\SI{1}{\ms}\ldots\SI{3}{\ms}$ \\
  DVS adapter time-step   & $\SI{1}{\ms}$ \\
  decoder time constant 
                            & $\SI{100}{\ms}$ \\ 
\end{params}

\bibliographystyle{plain}

\begin{thebibliography}{10}

\bibitem{bellec2017deep}
Guillaume Bellec, David Kappel, Wolfgang Maass, and Robert Legenstein.
\newblock Deep rewiring: Training very sparse deep networks.
\newblock {\em arXiv preprint arXiv:1711.05136}, 2017.

\bibitem{Bellec2018}
Guillaume Bellec, Darjan Salaj, Anand Subramoney, Robert Legenstein, and
  Wolfgang Maass.
\newblock Long short-term memory and learning-to-learn in networks of spiking
  neurons.
\newblock In {\em Conference on Neural Information Processing Systems (NIPS)},
  03 2018.

\bibitem{bengio2015towards}
Yoshua Bengio, Dong-Hyun Lee, Jorg Bornschein, Thomas Mesnard, and Zhouhan Lin.
\newblock Towards biologically plausible deep learning.
\newblock {\em arXiv preprint arXiv:1502.04156}, 2015.

\bibitem{bing2018end}
Zhenshan Bing, Claus Meschede, Kai Huang, Guang Chen, Florian Rohrbein, Mahmoud
  Akl, and Alois Knoll.
\newblock End to end learning of spiking neural network based on r-stdp for a
  lane keeping vehicle.
\newblock In {\em 2018 IEEE International Conference on Robotics and Automation
  (ICRA)}, pages 1--8. IEEE, 2018.

\bibitem{bing2018survey}
Zhenshan Bing, Claus Meschede, Florian Röhrbein, Kai Huang, and Alois~C.
  Knoll.
\newblock A survey of robotics control based on learning-inspired spiking
  neural networks.
\newblock {\em Frontiers in Neurorobotics}, 12:35, 2018.

\bibitem{dauce2009model}
Emmanuel Dauc{\'e}.
\newblock A model of neuronal specialization using hebbian policy-gradient with
  “slow” noise.
\newblock In {\em International Conference on Artificial Neural Networks},
  pages 218--228. Springer, 2009.

\bibitem{Djurfeldt2010}
Mikael Djurfeldt, Johannes Hjorth, Jochen~M. Eppler, Niraj Dudani, Moritz
  Helias, Tobias~C. Potjans, Upinder~S. Bhalla, Markus Diesmann, Jeanette
  {Hellgren Kotaleski}, and {\"{O}}rjan Ekeberg.
\newblock {Run-Time Interoperability Between Neuronal Network Simulators Based
  on the MUSIC Framework}.
\newblock {\em Neuroinformatics}, 8(1):43--60, mar 2010.

\bibitem{Ekeberg2008}
\"Orjan Ekeberg and Mikael Djurfeldt.
\newblock {MUSIC – Multisimulation Coordinator: Request For Comments}.
\newblock 2008.

\bibitem{falotico2017connecting}
Egidio Falotico, Lorenzo Vannucci, Alessandro Ambrosano, Ugo Albanese, Stefan
  Ulbrich, Juan~Camilo Vasquez~Tieck, Georg Hinkel, Jacques Kaiser, Igor Peric,
  Oliver Denninger, et~al.
\newblock Connecting artificial brains to robots in a comprehensive simulation
  framework: the neurorobotics platform.
\newblock {\em Frontiers in neurorobotics}, 11:2, 2017.

\bibitem{florian2007reinforcement}
R{\u{a}}zvan~V Florian.
\newblock Reinforcement learning through modulation of spike-timing-dependent
  synaptic plasticity.
\newblock {\em Neural Computation}, 19(6):1468--1502, 2007.

\bibitem{frey1997synaptic}
Uwe Frey, Richard~GM Morris, et~al.
\newblock Synaptic tagging and long-term potentiation.
\newblock {\em Nature}, 385(6616):533--536, 1997.

\bibitem{nest}
Marc-Oliver Gewaltig and Markus Diesmann.
\newblock Nest (neural simulation tool).
\newblock {\em Scholarpedia}, 2(4):1430, 2007.

\bibitem{gilra2017predicting}
Aditya Gilra and Wulfram Gerstner.
\newblock Predicting non-linear dynamics by stable local learning in a
  recurrent spiking neural network.
\newblock {\em Elife}, 6:e28295, 2017.

\bibitem{gilra2017nonlinear}
Aditya Gilra and Wulfram Gerstner.
\newblock Non-linear motor control by local learning in spiking neural
  networks.
\newblock In Jennifer Dy and Andreas Krause, editors, {\em Proceedings of the
  35th International Conference on Machine Learning}, volume~80 of {\em
  Proceedings of Machine Learning Research}, pages 1773--1782,
  Stockholmsmässan, Stockholm Sweden, 10--15 Jul 2018. PMLR.

\bibitem{hamid2016mesolimbic}
Arif~A Hamid, Jeffrey~R Pettibone, Omar~S Mabrouk, Vaughn~L Hetrick, Robert
  Schmidt, Caitlin~M Vander~Weele, Robert~T Kennedy, Brandon~J Aragona, and
  Joshua~D Berke.
\newblock Mesolimbic dopamine signals the value of work.
\newblock {\em Nature neuroscience}, 19(1):117, 2016.

\bibitem{izhikevich2007solving}
Eugene~M Izhikevich.
\newblock Solving the distal reward problem through linkage of stdp and
  dopamine signaling.
\newblock {\em Cerebral cortex}, 17(10):2443--2452, 2007.

\bibitem{kaiser2018synaptic}
Jacques Kaiser, Hesham Mostafa, and Emre Neftci.
\newblock Synaptic plasticity dynamics for deep continuous local learning.
\newblock {\em arXiv preprint arXiv:1811.10766}, 2018.

\bibitem{kaiser2016towards}
Jacques Kaiser, J~Camilo~Vasquez Tieck, Christian Hubschneider, Peter Wolf,
  Michael Weber, Michael Hoff, Alexander Friedrich, Konrad Wojtasik, Arne
  Roennau, Ralf Kohlhaas, et~al.
\newblock Towards a framework for end-to-end control of a simulated vehicle
  with spiking neural networks.
\newblock In {\em 2016 IEEE International Conference on Simulation, Modeling,
  and Programming for Autonomous Robots (SIMPAR)}, pages 127--134. IEEE, 2016.

\bibitem{Kappel2015}
David Kappel, Stefan Habenschuss, Robert Legenstein, and Wolfgang Maass.
\newblock {Network Plasticity as Bayesian Inference}.
\newblock {\em PLOS Computational Biology}, 11(11):e1004485, nov 2015.

\bibitem{spore}
David Kappel, Michael Hoff, and Anand Subramoney.
\newblock {IGITUGraz/spore-nest-module: SPORE version 2.14.0}.
\newblock Nov 2017.

\bibitem{Kappel2018}
David Kappel, Robert Legenstein, Stefan Habenschuss, Michael Hsieh, and
  Wolfgang Maass.
\newblock {A Dynamic Connectome Supports the Emergence of Stable Computational
  Function of Neural Circuits through Reward-Based Learning}.
\newblock {\em eneuro}, pages ENEURO.0301--17.2018, apr 2018.

\bibitem{Kappel2014}
David Kappel, Bernhard Nessler, and Wolfgang Maass.
\newblock {STDP Installs in Winner-Take-All Circuits an Online Approximation to
  Hidden Markov Model Learning}.
\newblock {\em PLoS Computational Biology}, 10(3):e1003511, mar 2014.

\bibitem{Kingma_Ba14_adammeth}
Diederik~P Kingma and Jimmy Ba.
\newblock Adam: A method for stochastic optimization.
\newblock {\em arXiv preprint arXiv:1412.6980}, 2014.

\bibitem{gazebo}
Nathan Koenig and Andrew Howard.
\newblock Design and use paradigms for gazebo, an open-source multi-robot
  simulator.
\newblock In {\em 2004 IEEE/RSJ International Conference on Intelligent Robots
  and Systems (IROS)(IEEE Cat. No. 04CH37566)}, volume~3, pages 2149--2154.
  IEEE, 2004.

\bibitem{Kruger2013}
Norbert Kruger, Peter Janssen, Sinan Kalkan, Markus Lappe, Ales Leonardis,
  Justus Piater, Antonio~J. Rodriguez-Sanchez, and Laurenz Wiskott.
\newblock {Deep hierarchies in the primate visual cortex: What can we learn for
  computer vision?}
\newblock 35(8):1847--1871, 2013.

\bibitem{legenstein2008learning}
Robert Legenstein, Dejan Pecevski, and Wolfgang Maass.
\newblock A learning theory for reward-modulated spike-timing-dependent
  plasticity with application to biofeedback.
\newblock {\em PLOS Computational Biology}, 4(10):1--27, 10 2008.

\bibitem{Lichtsteiner2008}
Patrick Lichtsteiner, Christoph Posch, and Tobi Delbruck.
\newblock {A 128$\times$128 120 dB 15 $\mu$s Latency Asynchronous Temporal
  Contrast Vision Sensor}.
\newblock {\em IEEE Journal of Solid-State Circuits}, 43(2):566--576, 2008.

\bibitem{lillicrap2015continuous}
Timothy~P Lillicrap, Jonathan~J Hunt, Alexander Pritzel, Nicolas Heess, Tom
  Erez, Yuval Tassa, David Silver, and Daan Wierstra.
\newblock Continuous control with deep reinforcement learning.
\newblock {\em arXiv preprint arXiv:1509.02971}, 2015.

\bibitem{mnih2016asynchronous}
Volodymyr Mnih, Adria~Puigdomenech Badia, Mehdi Mirza, Alex Graves, Timothy
  Lillicrap, Tim Harley, David Silver, and Koray Kavukcuoglu.
\newblock Asynchronous methods for deep reinforcement learning.
\newblock In {\em International conference on machine learning}, pages
  1928--1937, 2016.

\bibitem{mnih2015human}
Volodymyr Mnih, Koray Kavukcuoglu, David Silver, Andrei~A Rusu, Joel Veness,
  Marc~G Bellemare, Alex Graves, Martin Riedmiller, Andreas~K Fidjeland, Georg
  Ostrovski, et~al.
\newblock Human-level control through deep reinforcement learning.
\newblock {\em Nature}, 518(7540):529, 2015.

\bibitem{nakano2015spiking}
Takashi Nakano, Makoto Otsuka, Junichiro Yoshimoto, and Kenji Doya.
\newblock A spiking neural network model of model-free reinforcement learning
  with high-dimensional sensory input and perceptual ambiguity.
\newblock {\em PloS one}, 10(3):e0115620, 2015.

\bibitem{Neftci17_stocsyna}
Emre Neftci.
\newblock Stochastic synapses as resource for efficient deep learning machines.
\newblock In {\em Electron Devices Meeting (IEDM), 2017 IEEE International},
  pages 11--1. IEEE, 2017.

\bibitem{otsuka2010free}
Makoto Otsuka, Junichiro Yoshimoto, and Kenji Doya.
\newblock Free-energy-based reinforcement learning in a partially observable
  environment.
\newblock In {\em ESANN}, 2010.

\bibitem{pan2005dopamine}
Wei-Xing Pan, Robert Schmidt, Jeffery~R Wickens, and Brian~I Hyland.
\newblock Dopamine cells respond to predicted events during classical
  conditioning: evidence for eligibility traces in the reward-learning network.
\newblock {\em Journal of Neuroscience}, 25(26):6235--6242, 2005.

\bibitem{Pfister_etal06_optispik}
Jean-Pascal Pfister, Taro Toyoizumi, David Barber, and Wulfram Gerstner.
\newblock Optimal spike-timing-dependent plasticity for precise action
  potential firing in supervised learning.
\newblock {\em Neural computation}, 18(6):1318--1348, 2006.

\bibitem{ros}
Morgan Quigley, Ken Conley, Brian Gerkey, Josh Faust, Tully Foote, Jeremy
  Leibs, Rob Wheeler, and Andrew~Y Ng.
\newblock Ros: an open-source robot operating system.
\newblock In {\em ICRA workshop on open source software}, volume~3, page~5.
  Kobe, Japan, 2009.

\bibitem{rangel2016neurotransmitters}
Mauricio Rangel-Gomez and Martijn Meeter.
\newblock Neurotransmitters and novelty: a systematic review.
\newblock {\em Journal of psychopharmacology}, 30(1):3--12, 2016.

\bibitem{schulman2015trust}
John Schulman, Sergey Levine, Pieter Abbeel, Michael Jordan, and Philipp
  Moritz.
\newblock Trust region policy optimization.
\newblock In {\em International Conference on Machine Learning}, pages
  1889--1897, 2015.

\bibitem{Schulman2017}
John Schulman, Filip Wolski, Prafulla Dhariwal, Alec Radford, and Oleg Klimov.
\newblock Proximal policy optimization algorithms.
\newblock {\em arXiv preprint arXiv:1707.06347}, 2017.

\bibitem{tieck2018learning}
Juan Camilo~Vasquez Tieck, Marin~Vlastelica Pogan{\v{c}}i{\'c}, Jacques Kaiser,
  Arne Roennau, Marc-Oliver Gewaltig, and R{\"u}diger Dillmann.
\newblock Learning continuous muscle control for a multi-joint arm by extending
  proximal policy optimization with a liquid state machine.
\newblock In {\em International Conference on Artificial Neural Networks},
  pages 211--221. Springer, 2018.

\bibitem{Urbanczik_Senn14_learby}
Robert Urbanczik and Walter Senn.
\newblock Learning by the dendritic prediction of somatic spiking.
\newblock {\em Neuron}, 81(3):521--528, 2014.

\bibitem{Weidel2016}
Philipp Weidel, Mikael Djurfeldt, Renato~C. Duarte, and Abigail Morrison.
\newblock {Closed Loop Interactions between Spiking Neural Network and Robotic
  Simulators Based on MUSIC and ROS}.
\newblock {\em Frontiers in Neuroinformatics}, 10(31):1--19, aug 2016.

\bibitem{williams1992simple}
Ronald~J Williams.
\newblock Simple statistical gradient-following algorithms for connectionist
  reinforcement learning.
\newblock {\em Machine learning}, 8(3-4):229--256, 1992.

\bibitem{Wolf2017}
P.~Wolf, C.~Hubschneider, M.~Weber, A.~Bauer, J.~Härtl, F.~Dürr, and J.~M.
  Zöllner.
\newblock Learning how to drive in a real world simulation with deep
  q-networks.
\newblock In {\em 2017 IEEE Intelligent Vehicles Symposium (IV)}, pages
  244--250, June 2017.

\bibitem{Yu2016a}
Zhaofei Yu, David Kappel, Robert Legenstein, Sen Song, Feng Chen, and Wolfgang
  Maass.
\newblock Camkii activation supports reward-based neural network optimization
  through hamiltonian sampling.
\newblock {\em arXiv preprint arXiv:1606.00157}, 2016.

\bibitem{zenke2018superspike}
Friedemann Zenke and Surya Ganguli.
\newblock Superspike: Supervised learning in multilayer spiking neural
  networks.
\newblock {\em Neural computation}, 30(6):1514--1541, 2018.

\end{thebibliography}


\end{document}